# Neural cellular automata: applications to biology and beyond classical AI

Benedikt Hartl[1], Michael Levin[1,2], Léo Pio-Lopez[1],*

[1]Allen Discovery Center at Tufts University, Medford, MA, USA
[2]Wyss Institute for Biologically Inspired Engineering at Harvard University, Boston, MA, USA

* Author for Correspondence: leo.lopez@tufts.edu

**Abstract**

Neural Cellular Automata (NCA) represent a powerful framework for modeling biological self-organization, extending classical rule-based systems with trainable, differentiable (or evolvable) update rules that capture the adaptive self-regulatory dynamics of living matter. By embedding Artificial Neural Networks (ANNs) as local decision-making centers and interaction rules between localized agents, NCA can simulate processes across molecular, cellular, tissue, and system-level scales, offering a multiscale competency architecture perspective on evolution, development, regeneration, aging, morphogenesis, and robotic control. These models not only reproduce canonical, biologically inspired target patterns but also generalize to novel conditions, demonstrating robustness to perturbations and the capacity for open-ended adaptation and reasoning through embodiment. Given their immense success in recent developments, we here review current literature of NCAs that are relevant primarily for biological or bioengineering applications. Moreover, we emphasize that beyond biology, NCAs display robust and generalizing goal-directed dynamics without centralized control, e.g., in controlling or regenerating composite robotic morphologies or even on cutting-edge reasoning tasks such as ARC-AGI-1. In addition, the same principles of iterative state-refinement is reminiscent to modern generative Artificial Intelligence (AI), such as probabilistic diffusion models. Their governing self-regulatory behavior is constraint to fully localized interactions, yet their collective behavior scales into coordinated system-level outcomes. We thus argue that NCAs constitute a unifying computationally lean paradigm that not only bridges fundamental insights from multiscale biology with modern generative AI, but have the potential to design truly bio-inspired collective intelligence capable of hierarchical reasoning and control.

# 1 Introduction

Neural cellular automata (NCA) have recently emerged as versatile computational paradigm, bridging insights from biology and AI through their capacity to learn adaptive local rules that scale into distributed goal-directed behavior and system-level outcomes [1, 2]. Unlike traditional machine learning systems, which often struggle with robustness and generalization in open-ended, dynamic environments that require abstraction and reasoning [3-5], NCAs excel at modeling tasks ranging from evolutionary development and self-orchestrated morphogenesis [1, 6-8, 9 , 10-13], show great promise to better understand biological regeneration and aging [14-16], can efficiently be trained as robust distributed controllers of composite robotic entities [17-23], and are even competitive at challenging benchmarks for abstraction and reasoning tasks [24, 25] (see Figure 1). In this review, we survey modern trends in NCA research and literature, and relate their operational principles to the multiscale competency architecture of biological organization [26-30] (see Figure 1A and B), from multicellular collectives to hierarchical neuronal pathways in the neocortex [31-33]. NCAs show great promise as unifying computational framework for the emerging fields of diverse intelligence and truly bioinspired collective AI [34-36].

Cellular Automata (CAs) are one of the oldest algorithms we can find in the field of artificial life [37]. A cellular automaton is a discrete computational model consisting of a grid of cells, where each cell exists in one of a finite number of states, which evolve over discrete time steps based on a fixed deterministic rule. This rule is applied to all cells individually and will determine the state transition of the cells based on their own state and the one of their neighboring cells. The system evolves through discrete time steps and are dynamical systems which exhibit complex global behavior from simple local interaction and computation. Such a decentralized systems have been used to model the behavior of complex systems in biology and physics, and despite their simplicity, it has been shown that CAs like Conway's Game of Life, Elementary Cellular Automata, and Wireworld, are capable of universal computation [38, 39] showing their remarkable modeling capacity and representational richness. They have been used to model for example self-replication [40], universal constructors [41], biological pattern formation [42], complexity [43], and of most relevance here, they have been used to model emergent collective behavior in complex systems [44-48].

Very recently, CAs have gained a renewed attention in the context of artificial intelligence (AI). Modern computational capabilities have allowed researchers to develop more sophisticated CA models that can process information in a way very similar to neural networks [49]. Miller's pioneering work demonstrated the successful evolution of cellular automata with self-repair and self-regulation capabilities for producing the "French flag" pattern, establishing that evolutionary algorithms can identify multiple distinct rules sets capable of morphogenesis [50]. In the same perspective, Nichele et al. evolved compositional pattern producing networks for cellular automata morphogenesis and replication [42]. These developments have led to the emergence of NCA, which combine the distributed processing characteristics of traditional CA with the learning capabilities of neural networks [1, 2]. Unlike conventional CA with fixed rules, NCAs substitute their hardcoded local update functions with cell-specific Artificial Neural Networks (ANNs); as universal approximators they can be trained via gradient-based optimization or evolutionary algorithms on complex collective target behaviors to solve, for instance, the inverse task of target pattern assembly, or *in silico* morphogenesis.

Several important developments have expanded the scope and functionality of NCAs beyond basic morphogenesis tasks. Self-classification capabilities were demonstrated by Randazzo et al. [51] (see Figure 1F), who showed that NCAs could learn to categorize their own emergent patterns. Solving spatial problems like solving a maze were demonstrated too [12] (see Figure 1G). The concept of embedding spaces for morphogenetic processes was introduced through the NCA Manifold framework [52], enabling the representation of different developmental trajectories in structured latent spaces. Tesfaldet et al. incorporated self-attention mechanisms directly into the cellular update rules, allowing individual cells to selectively focus on relevant neighborhood features [53]. Information-theoretic principles were integrated into NCA design (2022), by introducing empowerment as an auxiliary objective function to encourage coordinated cellular behaviors that enhance system robustness and adaptability [54, 55]. The stochastic modeling of emergent dynamics was advanced by Palm et al. through their Variational Neural Cellular Automata framework, which employs variational inference to capture the probabilistic nature of pattern formation processes [56]. Multi-scale emergent phenomena were developed by Pande and Grattarola (2023), who proposed hierarchical NCA architectures capable of systematically modeling intercellular behaviors across different levels of resolution of hierarchically stacked NCAs [11].

The intersection of cellular automata with deep learning has opened new avenues for research in several domains. In computer vision, they are used for segmentation and texture synthesis applications [57-59]. In robotics, Variengien et al. investigated the evolutionary optimization of NCA-based control architectures for balancing tasks in cart-pole environments, demonstrating the feasibility of using distributed cellular computation for dynamic control problems [23]. And Horibe et al. explored the regenerative capabilities of NCAs in soft robotics applications, where cellular update rules enable autonomous morphological reconstruction following structural damage, similarly to the self-repair mechanisms observed in biological organisms [18, 19]. Pontes-Filho et al. demonstrated that growing NCAs can be used to first assemble an *in silico* robotic morphology, i.e., self-assemble a robotic body with tissue-, sensor-, and actuator-cell differentiation steps on an NCA's grid throughout a developmental timeframe, and second utilize the robots' composite body and the NCA's intercellular communication protocols as embodied policy for 2D navigation tasks [21] (see Figure 1H); relatedly, 3D self-assembly [60] and behavioral policies [22] (see Figure 1I) have been achieved. Furthermore, Najarro et al. demonstrated that 3D NCAs can be evolved to self-assemble hypernetworks that parametrize large-scale policy networks of RL agents, e.g., for the Lunar-Lander environment, an approach termed HyperNCA [20].

Recently, Hartl et al. successfully demonstrated that NCAs can be evolved to robustly control deformable microswimmers *in silico* in a fully decentralized way by utilizing locally coordinated body deformations in a physically realistic hydrodynamic environment; strikingly, these decentralized control policies generalized well across morphologies, i.e., swimmer sizes, and even allowed robust locomotion if large fractions of the swimmer's actuators were removed or disabled [17]. A related approach by Medvet et al. and Pigozzi et al. used local self-attention and NCA-like control paradigms to evolve embodied soft-robotic 3D morphologies [61, 62] which have assisted as *in silico* models to the successful creation of novel *in vitro* lifeforms, i.e., Xenobots [63] and Anthrobots [64, 65] (see Figure 1J): living, swimming self-powered biobots. Decentralized control in autonomous soft robotics thus holds great potential for future biomedical applications. NCAs have also been used to model complex biological phenomena, including for example morphogenesis [1, 22, 66], the scaling of goals from metabolic to organism-level goals via homeostasis [9], bioelectricity [67], memory [8], aging [14], cortex mechanisms [68], and evolution [7] (see section 3 for details).

Indeed, by design, the NCA architecture mirrors a large part of biological organization and particularly the multiscale competency architecture, where biological organisms exhibit nested hierarchies of goal-directed behaviors operating across different spatial and temporal scales [6, 26, 27, 36]. In biological systems, individual cells maintain homeostatic goals at the molecular level while simultaneously participating in tissue-level morphogenesis, organ-level functionality, scaling biological organization up to organ-level, organism-level behavior, and beyond. Similarly, NCAs show a multi-competency architecture by design as they are composed of cells embedding a neural network for decision-making and actions. Such NCAs can be organized into hierarchical structures where local cellular objectives emerge into progressively larger-scale system competencies [9, 11]. These systems can exhibit anatomical homeostasis [26] and, in turn, can serve as *in silico* model systems to study biological phenomena like aging and rejuvenation when the cybernetic tissues begin to decline [14]. Therefore, NCAs are models particularly suited to investigate multiscale information processing in biological or artificial life systems. In addition, the distributed architecture of NCAs show good robustness and error-correction mechanisms similarly to biological systems [9, 18]. Just as biological organisms can maintain functionality despite cellular damage or mutation through compensatory mechanisms, trained NCAs demonstrate remarkable resilience to perturbations, damage, or partial removal of cellular components while maintaining their ability to regenerate target patterns [1, 9, 18]. The temporal dynamics of NCAs also align closely with biological time scales, operating through iterative local updates that mirror the continuous, asynchronous cellular processes in living systems. Consequently, NCAs serve as a bridge between artificial intelligence and biology, offering a computational substrate that naturally incorporates the fundamental organizational principles underlying life itself. The interest of NCAs extends beyond pure biological applications and may present a new kind of AI based on collective intelligence and can show properties difficult to achieve in conventional AI and deep learning like self-repair capabilities, adaptive morphology and architecture, and the scaling of intelligence.

In this paper, we review the different use cases of NCAs for biological modelling, including morphogenesis and self-organization, aging, regeneration and bioelectricity, molecular design, and the generative genome. Then, we discuss why NCAs may be a model of choice for the modeling of biological systems based on the framework of biological multiscale competency architecture, i.e., by the idea that biological systems are made of agential material. We also discuss why they may be particularly important beyond classical AI, as NCAs are AI based on collective intelligence. We conclude with the challenges and limitations this kind of models are currently facing.

## 2 Neural cellular automata fundamentals

A CA comprises a discrete, typically 2D grid of cells $i \in 1, \ldots, N$, each maintaining a numerical, often discrete but potentially also continuous vector-valued state $s_i^t$. The state $s_i^t$ of each cell $i$ evolves over discrete time steps $t \to t+1$ via local transition rules $f: s_i^{t+1} \to s_i^{t+1}$ by considering the cells own state and all the states of its neighboring. Formally, this can be expressed as $s_i^{t+1} = f(s_i^t, \{s_j^t\}_{j \in \mathcal{N}(i)})$, where $\mathcal{N}(i)$ represents the neighborhood of cell $i$ on the CA's grid, often a Moore neighborhood. Even hardcoded update functions give rise to elaborate complex behaviors, but the framework allows any local update function (deterministic, stochastic, etc.).

NCAs extend CAs by using learnable – often differentiable – update rules $f \to f_\theta$ that are realized by an ANN with parameters $\theta$ and enable each cell i on the NCA to self-regularize its continuous vector valued states $x_i^t$ based on local perceptions of its local neighborhood $\mathcal{N}(i)$ on the grid. Importantly, in vanilla NCAs, all cells use the same ANN architecture for their updates and are thus symmetric in functionality, which nonetheless can give rise to asymmetric – and potentially incredibly rich – cell-state dynamics. In their most elemental form, NCAs implement convolutional neural networks (CNNs) with a single 3 × 3 convolutional layer followed by a dense feed forward layer that outputs a proposed cell-state update $\Delta x_i^t = f_\theta(x_i^t, \{x_j^t\}_{j \in \mathcal{N}(i)})$; the parameters $\theta$ thus describe synaptic weights and neuronal biases, typically amounting to a total number of ~10.000 [1] but depending on the application can range from $\approx 10^2$ [7, 69] to $\approx 10^6$ [70]. Formally, we can express such a state update as $x_i^{t+1} = x_i^t + \Delta x_i^t$.

In their seminal Growing NCA paper, Mordvintsev et al. demonstrated how to efficiently optimize the ANN parameters $\theta$ so the NCA will reliably self-assemble a 2D target image (e.g., of a lizard or smiley-face emojis) starting from a single seed cell on the grid [1] (see Figure 1C). This was realized by assigning the three items in each cell state vector as cell-specific RGB channels, $y_i^t \in \mathbb{R}^3$, which can be compared with the RGB values of the corresponding pixels of the target image, $\hat{y}_i$ (the full state $x_i^t$ can be higher-dimensional, e.g., 16D in [1]). Moreover, to realize real growth dynamics that starts from a single seed cell with (potentially trainable- [7, 8]) initial conditions $s_c^{t=0}$ in the center $c$ of the NCA's grid, a designated α (opacity) channel was introduced to distinguish "living" cells - with $\alpha > \alpha_T$ larger than a threshold of typically $\alpha_T = 0.1$ - from the background. In contrast, the states of "dead" cells are initially, and after every step set to zero. Only cells with one or more living neighbors can update their cell state and thus contribute to this *in silico* embryogenesis process. This makes the NCA's dynamics differentiable across temporal state updates $x_i^t \to x_i^{t+1}$ and allows employing gradient-based parameter optimization methods on global loss functions over emergent patterns $\{y_i^{t_D}\}_{i \in N}$ with respect to target patterns $\{\hat{y}_i\}_{i \in N}$. With this, the inverse problem of target pattern formation (of images or textures), but also novel applications such as self-regulated classification or image segmentation can be solved by means of differentiable collective intelligence.

To improve convergence on spatial self-assembly tasks, the NCAs cell states are often augmented with additional perception layers $\mathcal{P}(x_i^t, \{x_j^t\}_{j \in \mathcal{N}(i)})$ such as Sobel-like filters that extend cell state information with, e.g., local state gradients $x_i^t \to x_i^t \cup \mathcal{P}(x_i^t, \{x_j^t\}_{j \in \mathcal{N}(i)})$. To stabilize long-term behavior of the NCA's self-regulatory behavior, especially in morphogenesis tasks, it turned out effective to deliberately corrupt the NCA's cell states, e.g. by erasing (masking) parts of the pattern, and train the system to not only learn how to grow but also how to repair a target morphology. Such NCAs exhibit remarkable regeneration capabilities akin to axolots or planaria [1].

In non-differentiable situations or reinforcement learning (RL) applications, evolutionary algorithms or neuroevolution techniques have proven efficient to optimize NCA parameters [7, 9, 20, 21]. To increase robustness, stochastic cell updates have been introduced – via an update probability – that constrains the cells' reliability to regulate their own state by discarding proposed updates $\Delta x_i^t \to 0$ across stochastically chosen cells and time steps [1]. This explicitly implements an asynchronous update process across the NCA's grid that requires cells to generalize across cellular neighborhoods by learning how to distinguish between signal and noise. This avoids overfitting cellular policies on local cell-state dynamics and, in turn, promotes global patterns as collective attractor states enabling robust morphogenesis and self-repair. Notably, this renders NCAs as closely related to denoising diffusion models [71-75], as we and others discussed in detail elsewhere [6, 7, 70, 76] and below. Hartl et al. [7] improved both robustness and evolvability by utilizing noise in the update function $\Delta x_i^t \rightarrow \Delta x_i^t + \xi$; choosing or evolving the right combination of noise and cellular reliability (via the update probability) have proven effective to enhance learning and generalization capabilities of NCAs (see Figure 2A).

Although the NCA's grid stats serve as recurrent feedback signal in the unfolding dynamics – that gradually refines high-entropy initial conditions into structured samples conforming to the target data manifold – the majority of used ANNs are, thus far, feed-forward in nature. However, several contributions demonstrated that different ANN architectures, such as attention mechanisms[53], recurrent internal feedback layers [7], explicit separation of private and public cell states [8] (see Figure 1K), and even hierarchical NCA architectures with interconnected stacked layers [10], to name but a few examples, can not only be efficiently trained but scale the applicability of NCAs towards truly bio-inspired cybernetic systems, as we will discuss below.

NCAs can still be difficult to train, as solutions can collapse into trivial attractors (especially with gradient-based methods via pool-poisoning) or get trapped in suboptimal solutions that lack feature precision (predominantly in the case of neuroevolution training). They suffer from limitations in storage capacity (of multiple system-level target outcomes, or functional modalities) and in scaling; simulating realistic organ-level complexity at unicellular resolutions is currently infeasible. Relations of NCAs with reinforcement learning, active inference, or neuroevolution techniques remain underexplored. Moreover, their descriptive power, especially of biological systems, is still largely over-simplified: NCAs treat cells largely as homogeneous units and their numerical states and interactions are at best abstractions (or model composites) of most biological processes. And while progress has been made, it is difficult to integrate (and identify) NCA dynamics with molecular pathways, gene regulatory networks, or biomechanical forces. Training NCAs is mostly concerned with the accuracy of the final system-level outcomes, largely neglecting energetic, metabolic, or other physical/chemical/biological constraints necessary for developing viable morphologies; the learned update rules are still largely opaque black-boxes and difficult to interpret at the parameter and interaction level; update dynamics don't follow physical or biologically relevant ODEs; it is unclear how to implement reasonable multiscale coupling principles in HNCAs; in robotics, real-world experimental realization are in their infancies; and improving our theoretical understanding in stability, conditional dynamics, universality, criticality, and hierarchical phase-transitions still requires significant effort.

## 3 NCAs for AI-oriented computational biology

Modeling development and evolution with NCAs

In the last few years NCAs have been growingly applied in the simulation of morphogenesis [1, 7, 9-11, 22, 54]. Indeed, NCAs extend cellular automata with ANNs, replacing the typically hardcoded cell-state update rule for each cell, thereby providing multi-scale architectures particularly suited for the simulation of collectives of cells and tissues in morphogenesis. Very recently, the ENIGMA model enhanced the biological fidelity by replacing ANN with gene regulatory networks (GRNs) in each cell and incorporating cell-cell signaling pathways to mimic real biological signaling [77]. Evolutionary algorithms have been employed to ensure that ENIGMA can generate stable tissue patterns under varying conditions and thus draw a connection between computational models and biological systems. Similarly, [9] argued that biological systems can be understood as a hierarchy of homeostatic loops which range from cellular metabolic homeostasis to tissue and anatomical homeostasis. This concept is significant in drug design as in this view, molecules have to be targeted at higher levels than specific cellular processes or genetic targets, they also have to fit into the overall homeostatic networks that are trying to restore the setpoints of the body, like morphoceuticals [78]. NCAs have also been used to investigate the evolutionary implications of systems that can self-organize in a multi-competency architecture and found that the evolutionary process is greatly influenced by the cellular competency at several scales [7].

In these studies, NCAs have been employed to problems in morphogenesis, development and evolution, including the French flag problem, which is a model problem in the context of positional information in developmental biology. These simulations also demonstrated that simple competencies in cells can be scaled up to result in tissue-level organization and that there is robustness and error minimization in the formation of the anatomy [7, 9, 77]. Moreover, these systems demonstrate enhanced robustness against adversarial noise, generalizing well to changing system parameters, and exhibit enhanced transferability towards modified target patterning objectives [7, 22].

Very recently, EngramNCA has been introduced, a specialized NCA architecture that incorporates private, intracellular memory channels alongside publicly visible cell states, enabling decentralized memory storage and transfer while facilitating the emergence of hierarchical and coexisting morphologies [8]. Inspired by biological evidence of intracellular memory mechanisms beyond synaptic changes, EngramNCA allows cells to propagate "genetic" information through hidden channels, leading to self-organized growth of diverse structures from a common seed (see Figure 2B). This model demonstrates robust self-organization in morphogenesis, where simple local rules scale to produce complex tissue-like patterns with enhanced adaptability and regeneration capabilities, even in the presence of perturbations [8].

<u>Modeling regeneration and aging with NCAs</u>

One of the most promising applications of NCAs is in regenerative medicine. The paper "Growing Neural Cellular Automata" [1] has showcased the regenerative potential of NCAs by training such models to regrow patterns after simulated damage, such as cutting or removing sections of the target structure. By incorporating perturbations during the training process — including stochastic updates and pool-based sampling – these NCAs learn to persist and self-repair complex patterns like lizard or emoji shapes, mimicking biological regeneration where local rules enable recovery from unseen injuries [1]. Extending this to practical applications in soft robotics, researchers have developed NCAs that allow simulated robots to regenerate their morphology solely through local cell interactions, evolving separate automata for initial growth and post-damage repair [61, 62]. In experiments with 2D and 3D voxel-based robots, damaged morphologies (e.g., bipeds or tripods) achieved up to 99% structural similarity to originals, with varying locomotion recovery (67% in some cases), highlighting implications for resilient, self-organizing systems in dynamic environments [61, 62]. This regenerative framework has also been scaled to 3D artefacts, where NCAs generate and repair functional machines in virtual worlds like Minecraft, starting from a single cell. They demonstrate enhanced recovery (e.g., 99% block regeneration after halving a structure) when explicitly trained for damage resilience, paving the way for self-repairing physical structures inspired by biological morphogenesis [22]. More recent advancements integrate NCAs into diffusion models for parameter-efficient image synthesis, enabling efficient inpainting tasks where missing image regions are regenerated with high fidelity leveraging local communication to maintain contextual consistency and drawing analogies to biological tissue repair in fields like pathology imaging [70]. These examples underscore NCAs' versatility in modeling self-organized regeneration across scales, from cellular patterns to macroscopic structures.

NCA simulations have shown the important role of bioelectric patterns in guiding tissue development like in planarians and other organisms [79]. For example, NCAs were used to model disrupted craniofacial structures in tadpoles that subsequently reorganize and achieve correct anatomical configurations following a specific bioelectrical pattern, and provide a framework for therapeutic interventions [67].

NCAs have further been applied to aging, conceptualized as a decline in tissue morphology [14]. Evolutionary simulations showed that aging results from loss of goal directedness in morphospace at tissue level, leading to breakdown of collective anatomical homeostasis and morphological decline, even in the absence of genetic damage or accumulation of noise [14]. In addition, this study also revealed that there is dormant regenerative potential in aged tissues, which can be reactivated through targeted interventions. This finding reveals ways to develop strategies to rejuvenate tissues and extend healthy lifespan [14, 80].

Overall, the use of NCAs for modeling biology extends beyond morphogenesis. They can be used to understand aging and regeneration processes, and they provide insights into biological information processing at multiple levels—including genetic and bioelectrical. This modelling approach generates new hypotheses, sometimes experimentally validated [9, 67].

**NCAs as a model for the generative genome**

Evolution through developmental reproduction involves encoding the features and functionality necessary for high-fidelity reconstruction of an organism into the compact form of the genome; rather than directly encoding a "blueprint" or developmental "program", the genome more accurately represents a bowtie of latent variables instantiating organismal development. This process unfolds through gene regulatory mechanisms at the cellular level, informed by environmental cues and top-down feedback from higher organizational scales, establishing a multiscale framework [26, 30, 81]. Importantly, this process is neither deterministic nor simply emergent but can be related to a creative problem-solving layer between the genome and the phenotype of an organism that can deal with novel situations and environmental conditions on the fly. Such an architecture represents a form of collective intelligence from the bottom up and an embodiment of collective intelligence following William Jame's definition of intelligence "reaching the same goals by different means" [26, 27, 36].

The collective architecture of NCAs aligns closely with the nested bowtie structure of developmental biology, encompassing distributed gene regulatory mechanisms to emergent tissue or organ patterns. Each NCA cell maintains a numerical state akin to the physiological status of biological cells (e.g., transcriptomic expressions), while internal ANNs regulate localized cell states according to a collective system-level agenda (modelling biological cells' GRNs). Through that lens, the unfolding dynamics of NCAs – intercellular communication and intracellular information processing - closely models the distributed self-regulatory process of morphogenesis and morphostasis of distributed cellular agents interpreting and reacting to their local environment under novel morphogenetic contexts. The ANN of NCAs thus embody the generative genomic bottleneck for evolutionary development, not only positioning the genome as functional and generative unit, but empowering NCAs to serve as a distributed model for adaptively and creatively reconfiguring cellular expressions during the course of collective organismal development [6, 82].

Thus, NCAs might be excellent candidates for bridging the computationally irreducible gap between the agential layer of physiological computations connecting an organism's genotype and phenotype and *in silico* modeling and simulation via self-regulatory generative AI. While significant questions remain, NCAs may reflect real-life developmental pathways that can be intervened with in simulation, if trained on authentic biological data (as demonstrated through targeted adversarial reprogramming or takeover of tissue [15, 16], see Figure 1E). NCA research thus offers a transformative character in regenerative medicine, bioengineering, and synthetic biology [6].

**NCAs for medical applications: molecular design and interaction modeling**

NCAs have also been employed in drug discovery and protein docking. By representing small molecules and proteins as voxelized structures, NCAs have been used to design small molecule interactors, reconstruct protein structures, and model complex biomolecular interactions. These models are not only capable of simulating molecular docking and physical transformations but also of goal-directed molecule design by conditioning generation on specific protein targets. Moreover, NCAs may also be useful in reconstructing missing protein regions and modeling dynamical processes like isomerization, thus increasing their relevance to drug design and structural biology. However, the versatility of these NCAs models renders them somewhat inferior to the current state-of-the-art methods (such as diffusion- or transformer-based foundational models [83, 84], yet they exhibit promising potential in translating conceptual advances to practical applications in precision medicine and biomolecular engineering [85].

## 4 NCAs as a model of choice for biology, and beyond?

NCAs have an important characteristic that make them very suitable to address deep biological questions: their multiscale competency architecture. Indeed, biological systems show cognitive capabilities at different levels of biological complexity, from molecular networks to cells, tissue, organs until the whole organism and beyond (see Figure 1-A and B). This architecture enables distributed problem-solving where individual components possess limited competencies that collectively generate sophisticated system-level behaviors through dynamic interactions and information integration [26, 27]. In biology, we can find these competencies even below the cell level, GRNs have shown different types of learning and memories [86], and there is a large body of studies showing that we can find cognitive capabilities like memory, decision-making, anticipation, learning at all levels of biological complexity [26, 27]. At tissue-level, the collective behavior gives rise to wound healing, and morphogenesis that exceeds the capabilities of the individual agents [31]. At higher levels of organization, organs and the entire organisms integrate multiscale competency to achieve complex anatomical homeostasis and adaptive responses to the environment [87].

Evolution pivoted the same strategies used by cell collectives to navigate the space of anatomical possibilities during morphogenesis into brain-based navigation of 3D space, and then the emergence of higher intelligence [26, 27, 88]. Current AIs like LLMs, transformers etc. don't really leverage this multiscale competency architecture. Each formal neuron in a deep learning neural network, is treated only as a filter, firing with a predefined signal after a certain activation threshold is reached. These neuronal models are not capable of learning or forming a memory themselves; while learning and information processing dynamics only occur at the scale of the whole network. This might stand in stark contrast to biological neurons, especially to cortical columns in the human neocortex. The neocortex is formed by an integrated 2D grid of copies of the same neuronal circuit, the cortical columns, which have been argued to be capable of learning arbitrary concepts of our reality (objects, animals, other human beings, mental constructs, etc.). More importantly, these cortical columns might literally "model" the learned concepts and thus represent interactable reference frames or world models for relevant features of our Umwelt [32, 33, 89, 90]. Remembering would thus trigger models of past experiences, and thinking would translate not only into a navigation process through an associative conceptual space, but dynamically construct a network of interacting world models that are relevant under a certain context [6, 32, 33, 91]. NCAs might be excellent candidates to model such an architecture: they maintain a trainable ANN in each cell of an integrated grid which are in principle capable of representing reference frames of arbitrary concepts [92]. In turn, NCAs might not only be excellent models for biological self-organization as a multi-scale competency architecture, but even for higher-level cognitive processes of the human neocortex such as active perception [68, 93], raising fascinating questions about the parallels of morphogenesis and cognition [26, 27, 94].

Currently, the focus for scaling intelligence is based on the scaling of training data, model parameters, and computational resources [95]. This approach is incredibly resource intensive, and so-far mimics but a fraction of the capabilities of what biology achieves seemingly effortlessly. Several prominent AI researchers criticized the costs associated with exploiting scaling laws for commercializing large language models (LLMs) and ask for the development of new AI architectures beyond transformers. While LeCun promotes a world-model based approach [96], NCA architectures and their self-regulatory dynamics during inference represent a potential solution to this architectural gap by implementing truly biologically inspired multiscale competency architecture. Indeed, biological evolution modularly repurposed minimal collective goal-directed behavior (homeostatic loops) into ever higher-level problem-solving agents (composed of competent sub-agents) through competency amplification and an internal motivation for novelty, rather than data accumulation [9, 27]. By design, NCAs architecture implements the scaling of intelligence from local cellular communication pathways to system-level behavior and potentially to integrated world models.

Recently, ARC-NCA has demonstrated impressive performance on the abstract reasoning corpus (ARC) challenge, a collection of carefully designed tasks to benchmark modern AI frameworks for "general intelligence" and abstraction reasoning capabilities [4, 5]. Thus far, ARC remains an extremely challenging fundamental test for artificial general intelligence (AGI) both for explicitly crafted models as well as the most competitive large reasoning models, such as GPTs [3]. Interestingly, vanilla NCAs and EngramNCAs perform competitively on the ARC challenge compared to ChatGPT-4.5 – notably at the fraction of costs and computational resources – by leveraging developmental processes of simple interactive cellular agents that collectively self-assemble, *i.e.*, "grow", the complex patterns of the ARC-AGI-1 benchmark [24, 25]. In a sense, ARC-NCA models represent an embodiment of the problem-solving capacities necessary for the ARC challenge, reflecting the immense computational power that roots in embodied basal cognition [27, 34 ]. In that way, ARC-NCA can learn to generalize to grow the target-pattern modalities of many ARC-AGI-1 examples from a minimal set of two or three training examples, something standard AI and ML techniques struggle to achieve.

This raises fundamental questions about the organizing principles underlying biology's nested multiscale competency architecture: are composite agential objects embodied models of environmental features relevant at their scale [97]? How does this hierarchical representational structure emerge across spatial and temporal scales – from molecular machines to social collectives [9, 26]? Viewed through this lens, biological agents at every level function not merely as components or effectors, but as embodied representations of their respective environment, while pursuing developmental goals and systemic integrity. For example, a bistable enzyme may be an embodied model of ligand-specific decision-making; its conformational switch encodes a dynamic inference about molecular context. The transcriptome functions may be seen as an embodied intracellular model of cell-fate specification within the morphogenetic field. Bioelectric fields – coordinated by collective cellular behavior – serve as coarse-grained, spatiotemporally models guiding embodying cellular collectives toward large-scale pattern formation of target system-level outcomes [98]. Even animals can be understood as living world models [96, 99 ]: while their bodies are evolutionary fitted for the challenges of their Umwelt, their brains do not simply simulate the environment but embody its structure through continuous interaction. Neural circuits are not merely predictive tools, they function as dynamic, embodied representations of environmental regularities, integrating perception, action, and anticipation in real time [30, 32, 33, 91, 100]; this might even scale to swarms or groups as distributed models of environmental stability and multi-agential coordination [26, 28-30, 36, 101-106].

Crucially, these modeling aspects are not invented anew at every scale, but are regularities themselves – and if this process applies at every scale, this necessarily includes repurposing lower-level organizational units as building blocks to form higher-level embodied models [101, 107]. Evolution is tinkering, i.e., life does not design from scratch, it reuses, refactors, and repurposes across scales [26, 108]. Biological systems are not just composites of passive parts but are layers within layers of repurposable agential substrates – each unit simultaneously acting as an agent, a model, but also as signal. Proteins become signals in GRNs; the GRN state informs cell identity; cellular states coalesce into tissue-scale morphogenetic patterns, etc., all structured by self-modeling dynamics [6, 9, 26, 28, 29, 86, 109, 110]. Thus, biological organization might fundamentally be rooted in self-modelling of scale-dependent (relevant) environmental contexts through repurposable embodiments – scale-bridging representations that allow nested agents to maintain their physiological integrity against diffusive, entropic processes; this has been formalized on thermodynamic grounds via the variational free energy principle, or active inference [100]. This suggests that biology might be understood as models-within-models rather than merely layers-within-layers of organization – a multi-scale competency architecture where agency and embodied representation are inseparable. NCAs provide a computational framework that mirrors this hierarchical representational architecture: flexible local rules allow to generate global structure through embodied learning and coordination, offering a tractable way to explore how scale-bridging models emerge from simple, repurposable units and providing insight into the dynamics of nested biological self-modeling.

Recent developments by Béna et al. investigate the potential of training NCAs with gradient-based methods to perform universal computation in continuous state spaces. This was achieved by training NCAs with task-specific objectives to guide the emergence of fundamental computational primitives, i.e., logic gates which can be integrated dynamically by the NCA's dynamics universal NACs [111]. Relatedly, Miotti et al. demonstrated that differentiable logic gates can be used to model update functions of single cells in discrete CAs; these Differentiable Logic CAs can be efficiently trained end-to-end on system-level objectives for CAs with a discrete state space DiffLogicCA [112]. These ideas represent first promising steps towards self-assembling universal computation that is based on collective and distributed intelligence, and akin to the operational properties of biological matter.

Future studies will show whether the dynamics in computational medium of NCAs will undergo a similar transition from multicellular communication pathways observed in basal tissue to scale-free activation patterns observed in neuronal networks. This transition might involve explicit hierarchical architectures [10, 11] (see Figure 2C), or graph-like organization comprising short and long-range connections as opposed to the fixed grid layout of current NCAs, and might bring about novel understandings of the multiplicity of the computational power of different type of biological matter. For instance, recent promising studies argue that even vanilla NCAs that are pretrained to exhibit critical dynamics in their cell-state expressions represent more efficient initial conditions for learning down-stream tasks [68, 93, 113, 114]. Criticality in the cortical brain networks is hypothesized to maximize computational capacity by sustaining long-range correlations, power-law distributed neuronal avalanches, and efficient signal propagation across scales. Such a near critical state offers the exact mixture of structured signaling and operational flexibility necessary for optimal learning and information processing, balancing sensitivity to input with stability in network dynamics [115-119]. It has been argued that criticality is a universal setpoint for brain functions [120]. However, cognition is a spectrum and does not seem to be restricted to neuronal dynamics, the latter are only speed optimized for agents navigating our 3D Umwelt but the governing bioelectric dynamics are much more ancient [26, 27, 28, 31, 79, 121]: understanding biology as a multiscale competency architecture increasingly blurs the line between collective biological coordination and cognitive emergence. Criticality may not be merely a dynamical regime but a functional scaffold enabling self-regulatory agency across scales (from ion channels to tissues to behavior).

On this view, the brain's critical state is not the pinnacle of evolution but a fundamental tunable threshold allowing decentralized systems -- whether neural or non-neural -- to maintain homeostatic plasticity while preserving structural memory. Not only minds, but also biological forms, and, in general, collaborative collectives are solutions to information problems, and critical dynamics provide the optimal "sweet spot" for exploration vs. stability in both cognitive inference and developmental patterning [28]. Emulating such dynamics with NCAs may not only improve learning efficiency but also reveal fundamental principles linking biological organization to computational resilience -- offering a bridge between the physics of collective phenomena and the emergence of cognition. Through this lens, NCAs are not just tools, but might be models of life's underlying (near-critical) computational logic [6]. All together, these approaches represent technologically feasible ways to tackle the emerging research direction of Diverse Intelligence [35] and the TAME framework [27], and new forms of AIs based on collective intelligence.

## 5 Challenges and limitations

The ANNs inside each cell in NCAs are definitely less complex than large fully connected neural networks. Still, balancing model complexity (of the single-cellular agents) and interpretability of the emerging dynamics is far from trivial. It is unclear how to interpret and best analyze the integrated dynamics of such systems that undergo distributed computation. NCAs, thus far, represent toy-models for biological organization and for self-regulated complex systems in general, and may thus serve as transparent model system - their complexity and capabilities can be tuned by design - to study corresponding network dynamics, if not the dynamics of the evolution of such network dynamics.

Moreover, training NCAs to grow multiple target patterns, and thus store multiple system-level attractor states in the cellular ANNs, proved challenging. Some approaches tackle this shortcoming by co-evolving a multitude of initial conditions alongside shared ANN parameters per cell [7]. EngramNCA circumvents this by introducing a two-fold developmental process of a private and public cell state that operates through different communication pathways [8]. Yet, extending the pattern-storage and cell-resolution capabilities of NCAs to compete with modern generative AI techniques, not to speak of biology, is an active field of research [70, 122].

This shortcoming is closely related to the fact that current NCAs don't interface well: different trainings may lead to the same system-level outcome but implemented by completely different microlevel dynamics. This limits modularity and compatibility across multiple NCAs that operate in the same environment but developed not necessarily compatible communication strategies, and thus limits the analogy to biology which all talks the same language of biochemistry. Neuroevolution techniques offer some insights into transferability and evolvability experiments of NCAs [7, 10, 11] and EngramNCA demonstrates that even more biologically inspired architectures – based on alternating public state updates and private gene transfer – can interface morphologies belonging to different seed "genes" [8]; in general more research is necessary to this end. Interestingly, cues from non-equilibrium physics, i.e., from criticality and phase transitions [68, 93, 113, 114] suggest that NCA that are pretrained to operate at a critical state might be good "initial states" to learn further down-stream tasks. Thus, regulating NCA dynamics to near-critical behavior might advance their capabilities for adaptability.

Increasing the capabilities of NCAs might even call for hybrid training paradigms, combining differentiable learning with evolutionary strategies. Differentiable learning in NCAs, i.e., via backpropagation, can be highly efficient to learn task-specific dynamics (such as target pattern formation) with high precision [1, 76]. Evolutionary algorithms, on the other hand, not only offer a much greater diversity of potential solutions but bring forth NCAs that are naturally highly robust against noise, corruptions, and show enhanced evolvability and transferability to novel problems [7, 9-11, 18, 19, 54, 55]. Moreover, different ANN architectures – feed forward, circuit based, attention, recurrent, world models, etc. – might also strongly influence what the NCA can learn. A promising approach to not only combine the advantages of different learning paradigms (gradient-based and evolution) might be the recent conditional diffusion-evolution approach. This approach offers unprecedented quality-diversity scores, and through conditional sampling even allows to bias different evolutionary lineages towards a multitude of target outcomes [82, 123].

As already discussed above, the integrated multi-scale nature of biological organization poses significant challenges from a modelling and computational perspective. Hierarchical NCAs [10, 11] try to overcome this obstacle by imposing a fixed architecture of hierarchically stacked NCA layers of varying resolution (lower levels implementing the dynamics, while successive higher levels exerting control over lower levels). It remains however unclear, how to best decouple different scales of organization, or how to identify sub-groups of cells at certain that group together under different contexts. NCAs could learn from hierarchical reasoning approaches [124], from other multiagent-based RL techniques such as [125-128] or the recently proposed TAG framework for multi-agent hierarchical RL [129]. Another promising architecture are self-supervised techniques such as Joint Embedding Predictive Architectures (JEPA) which are closely related to nested world model architectures [96].

Furthermore, it remains unclear how to map the synthetic cell states of NCAs to their biological counter parts, i.e., to physiological properties of cells such as transcriptomic, proteomic, or other omics data. Notably, this is a critical missing link limiting the predictive power of NCAs for biology, which, if resolved, would render NCAs as excellent first-principle model biological dynamics. Although first efforts are made in this direction for the morphogenesis of cellular clusters [130], the problem remains unresolved so far, especially for multiscale organisms.

Notably, NCAs exhibit self-regulatory denoising behavior reminiscent of diffusion models [71-75], cutting edge foundational models of modern generative AI (e.g., Stable Diffusion [131] for image synthesis or SORA [132] for video synthesis, to name but a few). Both frameworks iteratively transform high-entropy, noisy initial states $x_t$ into structured data $x_0$ by reversing a diffusive process. While diffusion models are explicitly trained to predict and remove noise $\epsilon_t$ from corrupted data $x_t$ conditioned on a time step $t$ – which encodes the scale of corruption $x_t = x_0 + \epsilon_t$ – they leverage this temporal parameter to orchestrate hierarchical refinement: coarse features emerge early, followed by fine-grained details through successive symmetry-breaking transitions in the data manifold [133-135]. This enables structured, multi-scale learning and reveals deep insights into the hierarchical organization of features and symmetries in the (generative process of) data. In contrast, NCAs lack explicit temporal feedback and supervision; they do not receive $t$ as input. Instead, the dynamics – if at all – implicitly encode a developmental time through collective recurrent state updates, effectively internalizing optimal developmental trajectories of the target system-level outcome via embodied learning. This raises a key distinction: while diffusion models condition on time to guide denoising hierarchically, NCAs must learn this temporal structure end-to-end from spatiotemporal patterns alone, a significant challenge in capturing hierarchical organization through collective interactions. In turn, future NCA architectures could benefit from hybrid designs, enabling them to emulate diffusion-like hierarchical denoising while preserving their strengths in distributed, spatially aware computation. Such models may not only improve generative quality but also reveal how biological systems achieve complex morphogenesis through internalized temporal inference and embodied cognition via self-modeling across scales.

## 6 Conclusion

In the context of AI-oriented computational biology, we see NCAs as a model of choice. They integrate the biological multiscale competency architecture, and their inherent architecture allows to model self-organization and spatiotemporal multiscale information-processing systems for a wide range of biological phenomena including higher cognition [6, 8, 9, 14, 24, 25, 54, 60, 114]. By using NCAs to simulate biological processes like morphogenesis or tissue regeneration, we can gain deeper insights into the underlying principles of these complex biological systems at different levels of organization and how cognition scales following the principles of basal cognition [26, 27].

This relationship creates a positive feedback loop. By using NCAs to understand information-processing in cellular collective systems, we can in turn apply these biological principles to design new AI architectures and algorithms. In addition, developing such bioinspired AIs may be beneficial for the problem of alignment. Indeed, we can build AIs that are designed to be more adaptable, resilient, collaborative, and integrate modularly, leveraging the distributed capabilities of NCAs. And in addition, we can use all the toolbox of neuroscience and psychology to understand and interact with them as they will share the same multiscale competency architecture we find in biological systems. This symbiotic relationship —using NCAs to understand biology, and using that understanding to create truly bioinspired AIs (and possibly more aligned ones)— and despite all the challenges and limitations mentioned above, offers an optimistic new paradigm for computational biology and the future of AI and computation.

**Acknowledgements**

We thank Tomika Gotch for assistance with the manuscript. M.L. gratefully acknowledge support of Astonishing Labs via a sponsored research agreement to Tufts University, and of Templeton World Charity Foundation (via grant TWCF0606).

**Conflict of Interests Disclosure**

The Levin lab has a sponsored research agreement with Astonishing Labs, a company seeking to impact the longevity and biomedicine of aging field.

# Figures

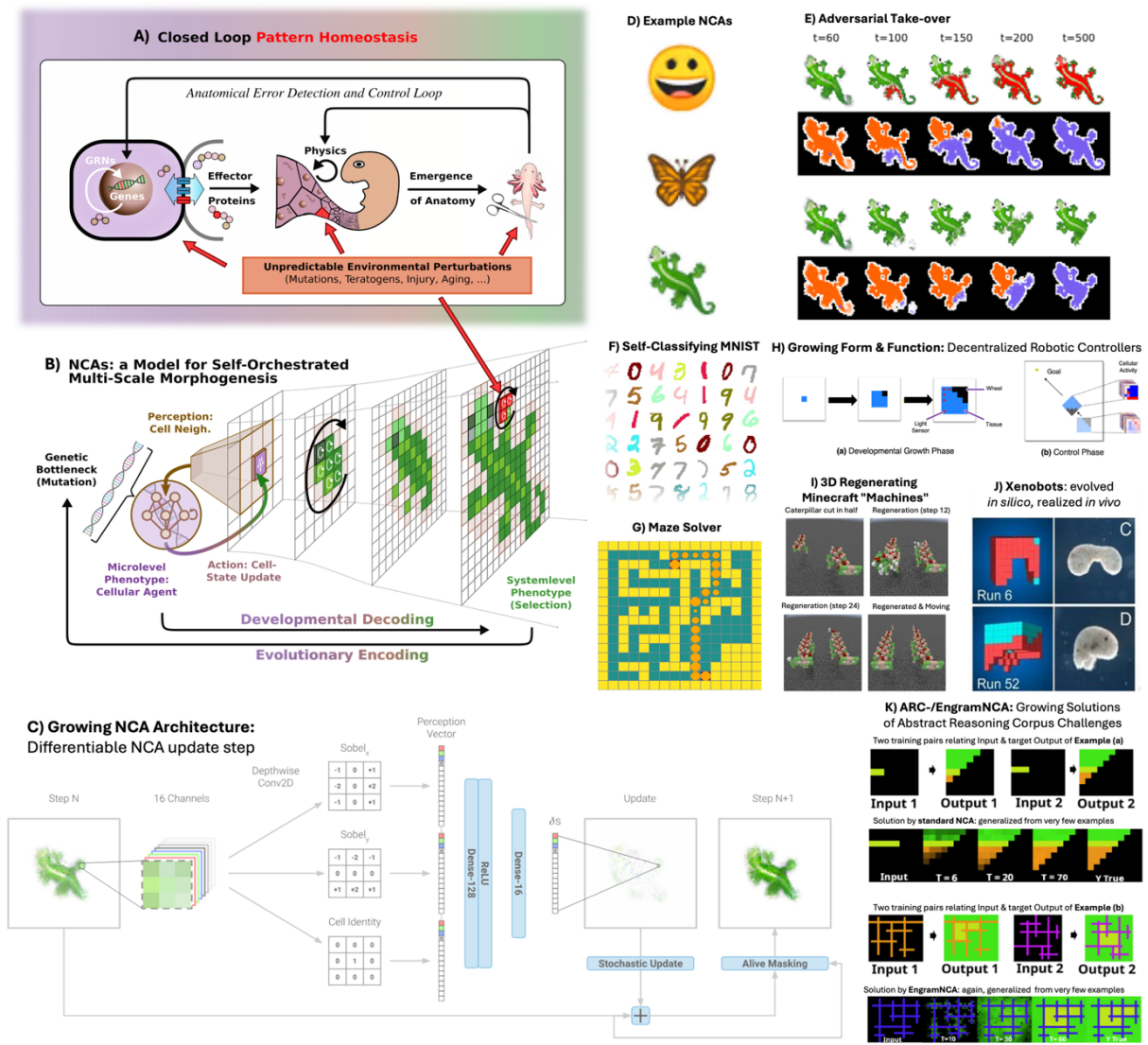

*Figure 1:* Overview of NCA models and applications. A) Development and morphogenesis as multiscale competency architecture via scale-free homeostatic error-correction [6] ; B) NCAs as models for biological evolutionary development through decentralized collaborative decision-making, and for the « generative genome » [6] ; C) Growing NCA architecture and self-regulatory update step applied on a 2D lizard morphology [1] ; D) Example patterns that can be learned by NCAs [1]; E) Adversarial take-over experiments of phenotypic traits such as skin color (top row) and phenotypic shape (bottom row) of pretrained NCAs [15, 16]; F) Self-classifying MNIST digits of NCA cells negotiating about what digit they are part of [51]; G) NCAs find shortest path and solve mazes [12]; H) NCAs as unifying substrate for growing form & function of decentralized robots [21] ; I) Self-Assembling and regenerating 3D Minecraft machines [22]; J) Xenobots – artificial, intelligent, yet fully biological novel lifeforms that have been designed in silico and realized in vitro [63]; K) EngramNCA trained and deployed on ARC-AGI-1 tasks [8, 24].

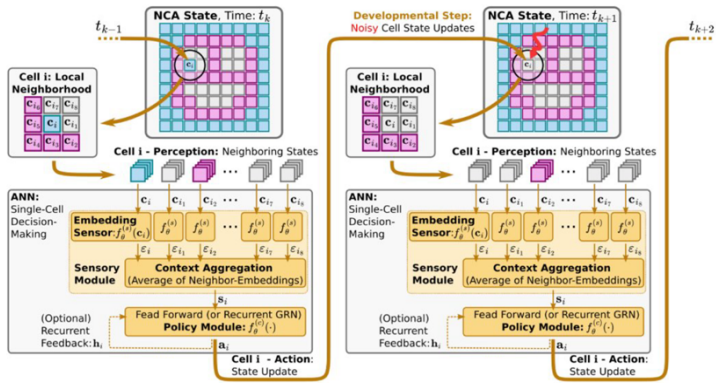

A) **Bio-inspired Recurrent Gene-Regulatory (RGRN)-like NCA:**
Perception via neighbor-specific embedding and permutation invariant context aggregation, with down-stream (hidden) RGRN policy-module; closely related to GNNs.

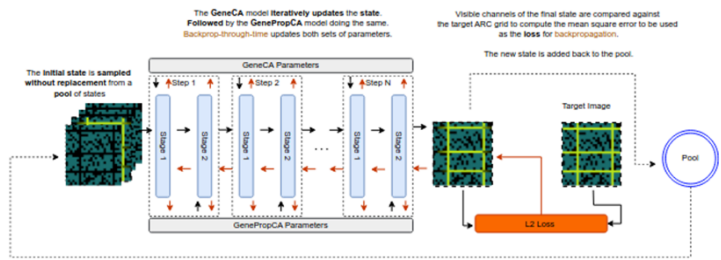

B) **EngramNCA:** GeneCA and GenePropCA are applied alternatingly on public and private cell states of the same CA grid, to grow ARC solutions conditionally to task-specific inputs.

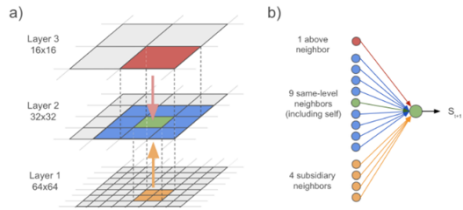

C) **Hierarchical NCA:** Several Layers of different resolution integrate into hierarchy of NCAs.

*Figure 2: Alternative Neural Cellular Automaton (NCA) Architectures. A) NCAs composed of cell-specific permutation invariant recurrent graph neural networks [7]: the Artificial Neural Network (ANN) of the NCA's cells are composed of an embedding module that filters and aggregates sensory input in a permutation invariant way into a context embedding, which is then further processed by a policy model – with potentially hidden recurrent states – that proposes cell state update actions in noisy environments. B) EngramNCA architecture with twofold update steps for public cell states via « GeneCA » and transfer of private genetic information via «GenePropCA » [8, 24]. C) Hierarchical NCAs (HNCAs), comprising multiple integrated (stacked) layer of NCAs of inreasingly coarse resolution at higher level in hierarchy. Cells integrate cell state information not only of their lateral neighbors, but of neighbors in adjacent layers as well. While lower level cells perform the most detailled dynamics, higher level NCAs progressively condition the collective behavior of adjacent lower levels and thereby can form increasingly more abstract representations and decisions. In that way, HNCAs develop hierarchal forms of control in multiscale architectures [10, 11].*